\documentclass[runningheads]{llncs}
\usepackage{graphicx}

\usepackage{tikz}
\usepackage{comment}
\usepackage{amsmath,amssymb} 
\usepackage{color}

\usepackage{booktabs}
\usepackage{multirow}
\usepackage{subcaption}
\usepackage[normalem]{ulem}
\useunder{\uline}{\ul}{}

\usepackage[pagebackref,breaklinks,colorlinks]{hyperref}

\newcommand\B[1]{\textcolor{blue}{#1}}

\usepackage[capitalize]{cleveref}

\newcommand{\figref}[1]{Fig.~\ref{#1}}

\newcommand{\tabref}[1]{Tab.~\ref{#1}}

\usepackage{enumitem}

\usepackage[accsupp]{axessibility}  

\begin{document}
\pagestyle{headings}
\mainmatter
\def\ECCVSubNumber{3086}  

\title{PCW-Net: Pyramid Combination and Warping Cost Volume for Stereo Matching} 

\titlerunning{Pyramid Combination and Warping Cost Volume for Stereo Matching}
%
\author{Zhelun Shen\inst{1}, 
Yuchao Dai\inst{2\dagger}, 
Xibin Song \inst{1\dagger}, 
Zhibo Rao \inst{2},
Dingfu Zhou\inst{1} \\ 
and Liangjun Zhang\inst{1} \\}
\authorrunning{Shen, Zhelun et al.}

%
\institute{Robotics and Autonomous Driving Lab, Baidu Research, China \and
Northwestern Polytechnical University \\
\email{\{shenzhelun, song.sducg, dingfuzhou\}@gmail.com;} 
\email{daiyuchao@nwpu.edu.cn; raoxi36@foxmail.com; zhangliangjun@baidu.com;}
}

\maketitle
\renewcommand{\thefootnote}{\fnsymbol{footnote}}
\footnotetext[4]{Corresponding authors}
\renewcommand{\thefootnote}{\arabic{footnote}}
\setcounter{footnote}{0}

\begin{abstract}
Existing deep learning based stereo matching methods either focus on achieving optimal performances on the target dataset while with poor generalization for other datasets or focus on handling the cross-domain generalization by suppressing the domain sensitive features which results in a significant sacrifice on the performance. To tackle these problems, we propose PCW-Net, a \textbf{P}yramid \textbf{C}ombination and \textbf{W}arping cost volume-based network to achieve good performance on both cross-domain generalization and stereo matching accuracy on various benchmarks. In particular, our PCW-Net is designed for two purposes. First, we construct combination volumes on the upper levels of the pyramid and develop a cost volume fusion module to integrate them for initial disparity estimation. Multi-scale receptive fields can be covered by fusing multi-scale combination volumes, thus, domain-invariant features can be extracted. Second, we construct the warping volume at the last level of the pyramid for disparity refinement. The proposed warping volume can narrow down the residue searching range from the initial disparity searching range to a fine-grained one, which can dramatically alleviate the difficulty of the network to find the correct residue in an unconstrained residue searching space. When training on synthetic datasets and generalizing to unseen real datasets, our method shows strong cross-domain generalization and outperforms existing state-of-the-arts with a large margin. After fine-tuning on the real datasets, our method ranks $1^{st}$ on KITTI 2012, $2^{nd}$ on KITTI 2015, and $1^{st}$ on the Argoverse among all published methods as of 7, March 2022. The code will be available at https://github.com/gallenszl/PCWNet. 
\end{abstract}


\section{Introduction}
\label{sec:intro}
Stereo matching aims to estimate the disparity map between a rectified image pair, which contributes to various applications, such as autonomous driving \cite{autonomousdriving} and robotics navigation \cite{roboticsnavigation}. Benefiting from the unprecedented development of deep learning technologies, remarkable progress has been achieved in solving the task of stereo matching.

\begin{figure}[!t]
	\centering
	\tabcolsep=0.02cm
	\begin{tabular}{c c}
	\includegraphics[width=0.35\linewidth]{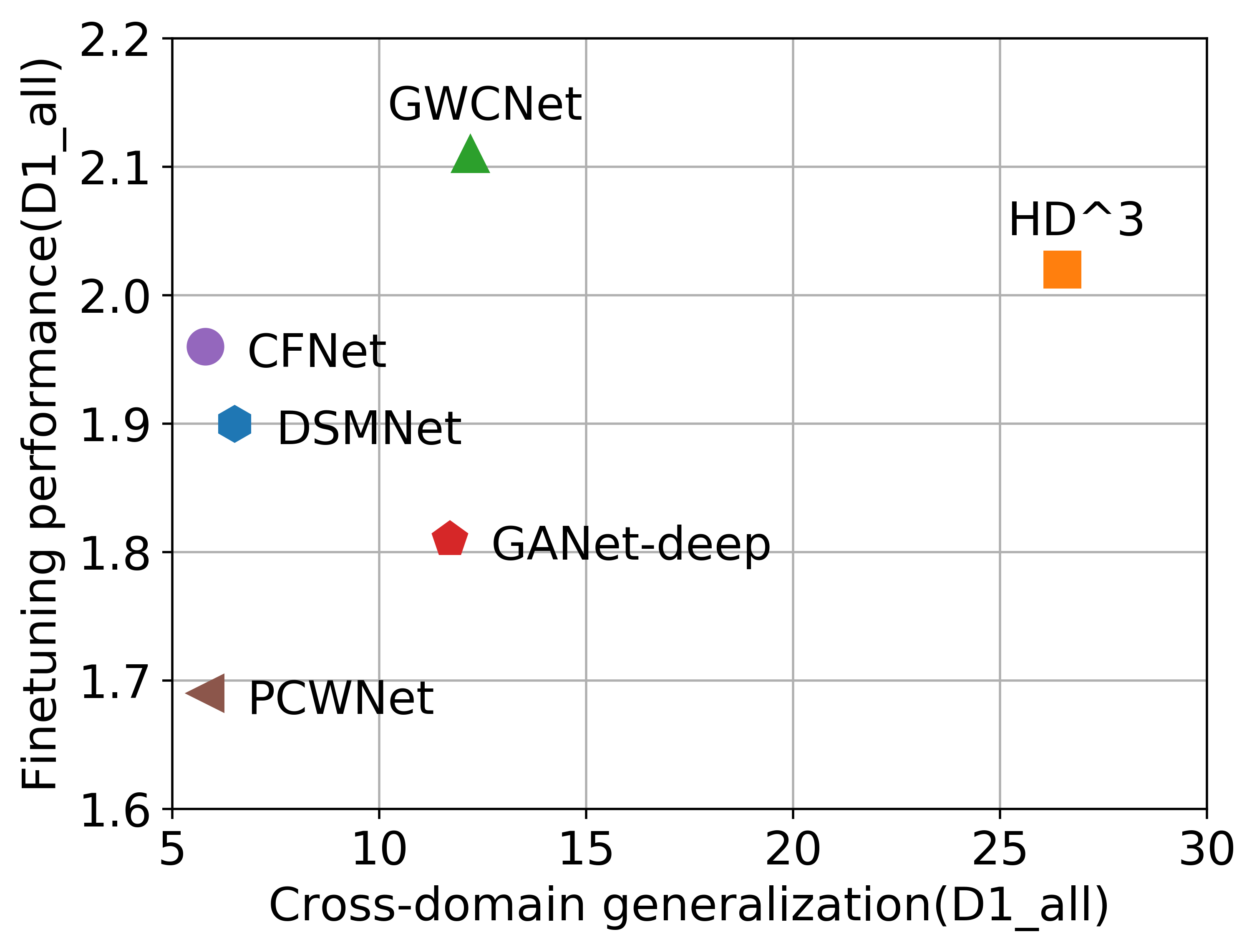}&
     \includegraphics[width=0.35\linewidth]{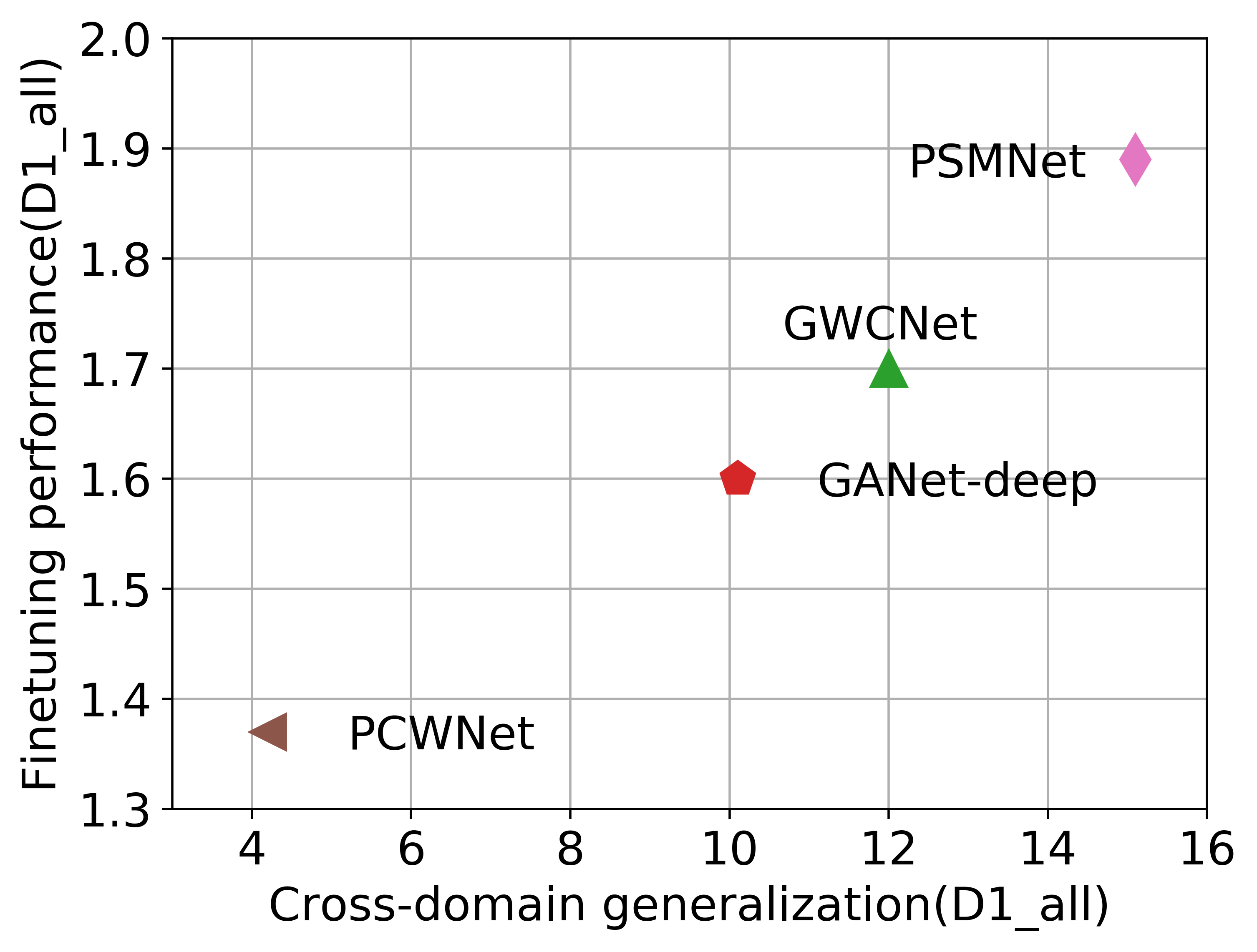}\\
	{(a) KITTI 2015} &	{(b) KITTI 2012 } \\ 
	\end{tabular}
	\vspace{-0.15in}
	\caption{\footnotesize \textbf{Model generalization ability vs fine-tuning performance on KITTI 2012\&2015 datasets.} X-axis: all methods are trained on synthetic datasets and then tested on KITTI training sets to evaluate the cross-domain generalization. Y-axis: all methods are finetuned on the KITTI training sets and then tested on KITTI testing sets to evaluate the fine-tuning performance. D1\_all is used for evaluation (the lower the better) and PCWNet is our method, which achieves the best overall performance.}
	\vspace{-0.25in}
	\label{fig: comparsion generalization and accuracy}
\end{figure}

	


To achieve remarkable stereo matching performance, approaches \cite{rao} are usually trained on large-scale synthetic datasets (e.g., SceneFlow \cite{dispnet}) first and then fine-tuned on limited target dataset collected from the real scenarios such as KITTI \cite{KITTI_2012}, Middlebury \cite{Middlebeurg_2002taxonomy}, and ETH3D \cite{eth3d}. By extracting representative features \cite{psmnet,emcua} and constructing powerful cost volume \cite{gcnet,gwcnet}, these methods achieve state-of-the-art performances on most of the standard stereo matching benchmarks. However, their performance decreases dramatically on unseen real-world scenes due to the large domain gaps across different datasets. Furthermore, these methods even cannot achieve consistent fine-tuning performances on different real-world datasets from similar scenarios. For example, some methods \cite{psmnet,ganet} perform well on the KITTI datasets \cite{KITTI_Stereo_2015,KITTI_2012}, while having limited performances on the Argoverse benchmark \cite{Argoverse_2_2021} with high image resolutions though both of them are collected by a driving vehicle in the traffic environment.


Meanwhile, many approaches~\cite{dsmnet,cfnet,improving} are also specifically designed to handle domain generalization issues in stereo matching which aims to improve the generalization of the network to unseen scenes. By incorporating geometry priors and extracting domain-invariant features, these methods show strong cross-domain generalization when trained on synthetic datasets and generalized to unseen real datasets. However, such methods \cite{dsmnet} normally need a significant sacrifice on accuracy to improve the cross-domain generalization due to the filtration of domain-sensitive features. Thus, a key problem for further research is designing a framework that can achieve excellent performances on the target dataset and also have satisfactory generalization ability to novel scenarios.

To relieve the issue mentioned above, we introduce the PCW-Net to construct a \textbf{P}yramid \textbf{C}ombination and \textbf{W}arping cost volume to hit two birds with one stone for achieving both generalization ability and good performance. Specifically, we use the pyramid cost volumes for two purposes. On one hand, we construct multi-scale combination volumes on the upper levels of the pyramid and develop a cost volume fusion module to integrate them for initial disparity estimation. The pyramid cost volume aims to cover multi-scale receptive fields and boost the network to see different scale regions of the original image. Thus, multi-level information can be fused together, i.e., textures, contours, and areas. Typically, non-local information (such as contours and area) is more robust to domain changes, thus better performance and generalization ability for different resolutions of images can be obtained. On the other hand, we also construct a 3D warping volume at the final level of the pyramid to further refine the initial disparity map. With the constructed 3D warping volume, we can narrow down the residue searching range from an initial disparity searching range to a fine-grained one, which can dramatically alleviate the difficulty of the network to find the correct residue in an unconstrained residue searching space.

To prove the effectiveness of the proposed PCW-Net, we perform extensive experimental evaluations on various benchmarks to verify its fine-tuning performance and generalization ability. When trained on synthetic datasets and generalized to unseen real-world datasets, PCW-Net shows strong cross-domain generalization and outperforms best prior work \cite{cfnet} by a noteworthy margin. After fine-tuning on the real dataset, our method can achieve consistent SOTA performance across diverse datasets. Specifically, it ranked first on KITTI 2012 leaderboard\footnote{\href{http://www.cvlibs.net/datasets/kitti/eval_stereo_flow.php?benchmark=stereo}{http://www.cvlibs.net/datasets/kitti}}, second on KITTI 2015 leaderboard, and first on Argoverse leaderboard\footnote{\href{https://eval.ai/web/challenges/challenge-page/917/leaderboard/2412}{https://eval.ai/web/challenges/challenge-page/917/leaderboard/2412}} \cite{Argoverse_2_2021} among all published methods as of 6 March 2021. As demonstrated in Fig.~\ref{fig: comparsion generalization and accuracy}, our method can achieve the best overall performance when considering both the fine-tuning accuracy and cross-domain generalization on the KITTI 2015 benchmark.

Our main contributions can be summarized as: \vspace{-0.2cm}
\begin{itemize}[leftmargin=*]
\item An effective framework, $i.e.$, PCW-Net, is proposed which achieves remarkable generalization ability from synthetic dataset to real dataset while also excellent performances on the various target benchmarks after model fine-tuning. 
\item A novel multi-scale cost volume fusion module is proposed to cover multi-scale receptive fields and extract domain-invariant structural cues, thus better stereo matching performance of different resolutions of images is achieved.
\item An efficient warping volume-based disparity refinement module is proposed to narrow down the unconstrained residue searching space to a fine-grained one, which can dramatically alleviate the difficulty of the network to find the correct residue in an unconstrained residue searching space.
\item The proposed PCW-Net set new SOTA performance on both KITTI 2012 and Argoverse leaderboards among all the methods with publications, while it also achieves the $2^{nd}$ on the KITTI 2015 benchmark.
\end{itemize}

\begin{figure*}[!t]
 \centering
 \includegraphics[width=\linewidth]{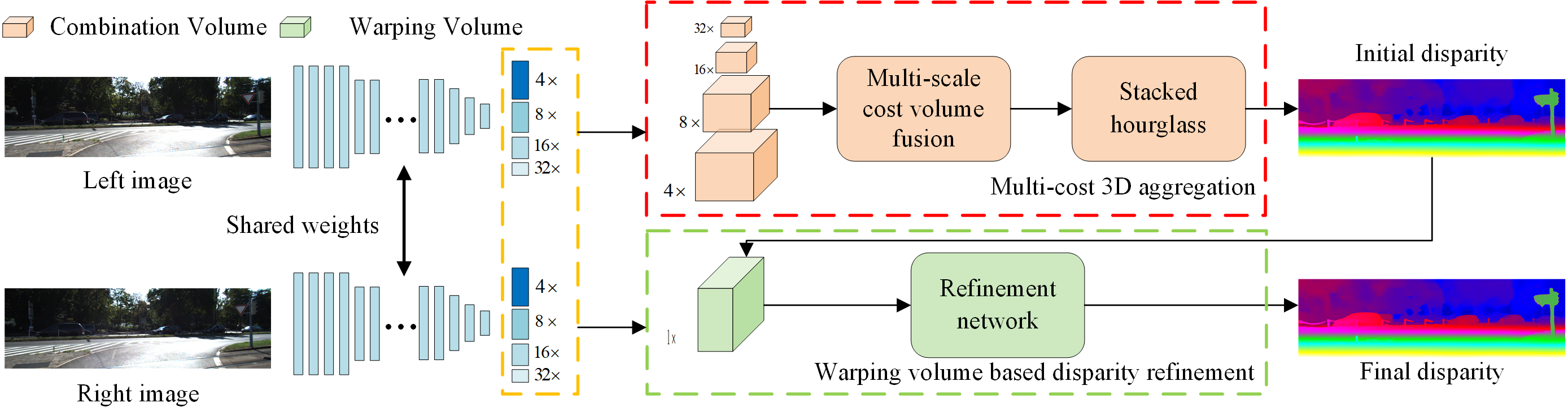} 
 \vspace{-0.3in}
 \caption{\footnotesize General Structure of the proposed PCW-Net, which consists of three main modules as multi-scale feature extraction, multi-scale combination volume based cost aggregation, and warping volume based disparity refinement.}
 \vspace{-0.25in}
 \label{fig: architecture}
\end{figure*}

\section{Related Work} \label{sec:related_work}

\noindent\textbf{Cost Volume based Deep Stereo Matching.} DispNet \cite{dispnet} first introduces the concept of cost volume (correlation volume) into end-to-end stereo matching methods. Following this work, GCNet \cite{gcnet} proposes to construct concatenation volume and regularize it with 3D convolution layers and GwcNet \cite{gwcnet} introduces group-wise correlation to provide better similarity measures. For all these prior works, cost volume construction has been placed in an extremely important position and deserves further exploration.

\noindent\textbf{Deep Stereo Matching with Disparity Refinement.} Recently, many researchers \cite{crl,madnet,pwcnet,mcvmsc,deeppruner,ednet,displacement} attempt to integrate the disparity refinement step into an end-to-end learning model. \cite{crl} introduces a two-stage network called CRL in which the first stage extends DispNet \cite{dispnet} to get an initial disparity map and the second stage refines the initial disparity map in a residual manner. Different from CRL, MCV-MFC \cite{mcvmsc} proposes to calculate reconstruction error in feature space rather than color space and share features between disparity estimation network and refinement network. PWCNet \cite{pwcnet} proposes a context network, which is based on dilated convolutions to refine flow. However, existing methods mainly depend on the fitting capabilities of the networks to directly regress a residue with context information. Different from these works\cite{crl,mcvmsc}, here we introduce the warping volume to guide the disparity refinement. Specifically, the warping volume is constructed by warped right image features and left image features according to a pre-defined residue range. That is the warping volume narrows down the residue searching space from initial disparity searching space to a fine-grained one, which makes the network easier to find the corresponding pixel-level residue.

\noindent\textbf{Multi-scale-based Deep Stereo Matching.} Multi-scale information has been widely employed in deep stereo matching methods.  These methods can be roughly categorized into two types: (1) The first category \cite{psmnet,mcvmsc,emcua} usually employs a multi-scale feature extraction network to generate feature maps at different scales and then fuse them to construct a single volume at a fixed resolution. That is these methods mainly use multi-scale features rather than multi-scale cost volumes. (2) The second category \cite{cvpmvsnet,cascade,uscnet,uasnet} proposes to construct cascade pyramid cost volume and progressively regress a high-quality disparity map from the coarsest cost volume. That is these methods employ each scale cost volume to estimate disparity maps separately. Different from the former two categories, our work selects to directly fuse multi-scale combination volumes to capture a more robust feature representation for initial disparity estimation. Then, we employ warping volume to further refine the initial disparity. More related to our work is SSPCV \cite{sspcv}, which also proposes a cost volume fusion module. However, SSPCV just fuses the pyramid cost volume by constantly employing 3D hourglass modules to regularize the upsampled cost volume. Such operation is time-consuming and GPU memory-unfriendly.

\section{Proposed Approach} \label{sec:approach}
We propose a PCW cost volume to effectively exploit the multi-scale cues for accurate and robust disparity estimation. The architecture of our network is illustrated in Fig. \ref{fig: architecture}, which consists of three parts: multi-scale feature extraction, multi-scale combination volume based cost aggregation, and warping volume based disparity refinement. Specifically, the extracted multi-scale features are first employed to construct a pyramid cost volume. Then, the pyramid volumes have been used for two purposes. Firstly, we construct combination volumes on the upper levels of the pyramid and develop a cost volume fusion module to integrate them for initial disparity estimation. Secondly, we construct the warping volume at the last level of the pyramid for disparity refinement. Detailed of each module will be introduced as follows.

\subsection{Multi-scale Features Extraction}

\begin{figure}[t!]
 \centering
 \includegraphics[width=0.98\linewidth]{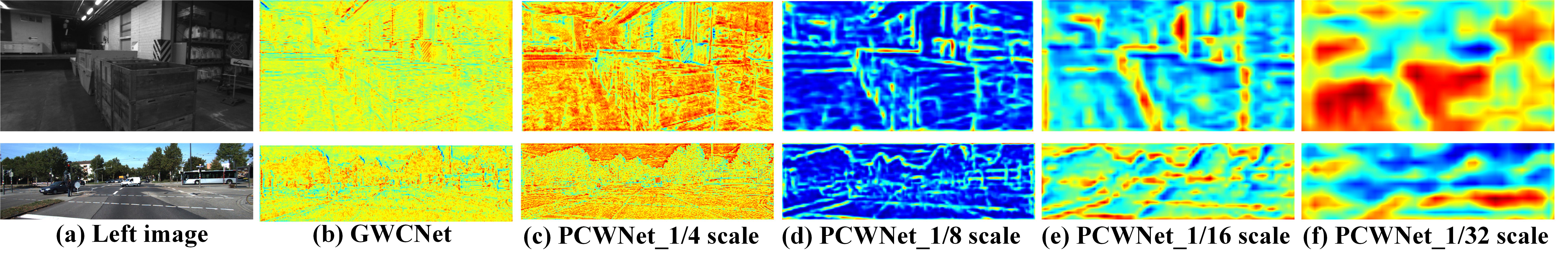} 
 \vspace{-0.15in}
 \caption{\small Visualization of extracted multi-scale feature maps on two real-world datasets (from top to bottom: ETH3D and KITTI). All methods are trained on synthetic data (SceneFlow) and tested on unseen real scenes. Note that GWCNet only extracts feature maps at 1/4 scale for following single-scale cost volume construction while our method extracts multi-scale feature maps for pyramid cost volume construction. More qualitative results are given in the supplementary materials. }
 \vspace{-0.2in}
 \label{fig: feature_visual}
\end{figure}

As shown in Fig.~\ref{fig: architecture}, given an image pair, following the Resnet-like network proposed in \cite{gwcnet,psmnet}, we use three convolution layers with ${\rm{3}} \times {\rm{3}}$ kernels, four basic residual blocks, and a $\times 2$ dilated block to get the unary feature map at the first level (1/4 of the original input image size). Then three residual blocks with stride $2$ are employed to obtain the feature maps at the other three levels with $\frac{1}{8}$, $\frac{1}{16}$ and $\frac{1}{32}$ of the original input image size. With the extracted features, a series of pyramid cost volumes can be constructed at different levels.


\subsection{Combination volume based 3D Aggregation}

We propose to construct multi-scale combination volumes and develop a cost volume fusion module for initial disparity estimation. Previous work \cite{dsmnet} observes that the limited effective receptive field of current deep stereo matching methods will drive the network to learn domain-sensitive local features. Instead, our method can cover multi-scale receptive fields and boost the network to see different scale regions of the original image by fusing multi-scale cost volumes. As shown in the figure \ref{fig: feature_visual}, we visualize the extracted multi-scale feature map on various of real datasets. It can be seen from figure \ref{fig: feature_visual}. (b) that GWCNet~\cite{gwcnet} only extracts 1/4 scale features of the input image, which only contains local information such as textures, thus the performance is limited. On the contrary, our method extracts features with multi-scales, which contains much more high-level information (sub-figs (c)-(f)), i.e., textures (c), contours (d,e), and areas (f). Typically, non-local information (such as contours and area) is more robust to domain changes and that is why our method achieves better generalization ability. Moreover, sub-figs (a) shows that the used two real datasets have significant domain shifts, e.g., indoors vs outdoors and color vs gray. However, our method can still extract domain-invariant contours (sub-figs (d)-(e)) and areas (sub-figs (f)) information from two real datasets, which further verifies the effectiveness of the proposed method. In addition, high-level information, i.e., contours and area can drive the network to better learn the affiliation between an object and its sub-region, e.g., textureless regions and repeated patterns such as car window is a part of the car, thus, better performance and generalization ability for different resolutions (high and low resolutions) of images can be obtained.


\subsubsection{Multi-scale Combination Volume Construction}

The combination volume is constructed at 4 pyramid levels and for each level $i$, the combination volume $V^{i}_{comb}$ is a 4D volume with the size of ($H^{i} \times W^{i} \times D^{i} \times C$) which includes concatenation volume $V^{i}_{concat}$ and group-wise correlation volume $V^{i}_{corr}$ \cite{gwcnet}, where ($H^{i}$, $W^{i}$) is the spatial size, $C$ is the combined feature dimension. Assuming the extracted feature at level $i$ is $f^{i}$, then the combination volume $V_{comb}^i$ can be computed as:
\vspace{-0.1in}
\begin{equation}
\begin{aligned} 
    V_{comb}^i & = V_{concat}^i\: || \:V_{corr}^i, \\
   V_{concat}^i(d,x,y) & = {\delta _1}(f_L^i(x, \:y))\: || \:{\delta _1}(f_R^i(x - d, \:y)) \\
   V_{corr}^i(d,x,y,g)  & = \frac{1}{{N_c^i/{N_g}}}\left\langle {{\delta _2}(f_{L}^{ig}(x, \:y)), {\delta _2}(\:f_{R}^{ig}(x - d, \:y))} \right\rangle \\
\end{aligned}
    \label{eq:volume_construction}
    \vspace{-0.1in}
\end{equation}
where $||$ denotes concatenation operation at the feature axis and $f^{i}_{L}$, $f^{i}_{R}$ are extracted features at left and right images respectively. $f^{ig}$ are grouped features, which are evenly divided from the extracted feature $f^{i}$ according to the number of group $N_g$.
$d$ denotes all disparity levels in $(0,D_{\max }^i)$, $N_c$ is the channels of $f^{i}$ and $\left\langle {{\rm{  }},{\rm{  }}} \right\rangle$ represents the inner product. Different with gwcnet\cite{gwcnet}, during the construction of combination volume, we add one more convolution layer without activation function and batch normalization (named as normalization layer $\delta$) to make the two terms of feature ($f^{i}$ and $f^{ig}$) share the same data distribution. Experimental results show that this simple while efficient operation can optimize the two terms of cost volume complementary to each other and thus promote the final performance. Then the multi-scale combination volume will be fused together to predict the initial disparity map.


\begin{figure}[t!]
 \centering
 \includegraphics[width=0.98\linewidth]{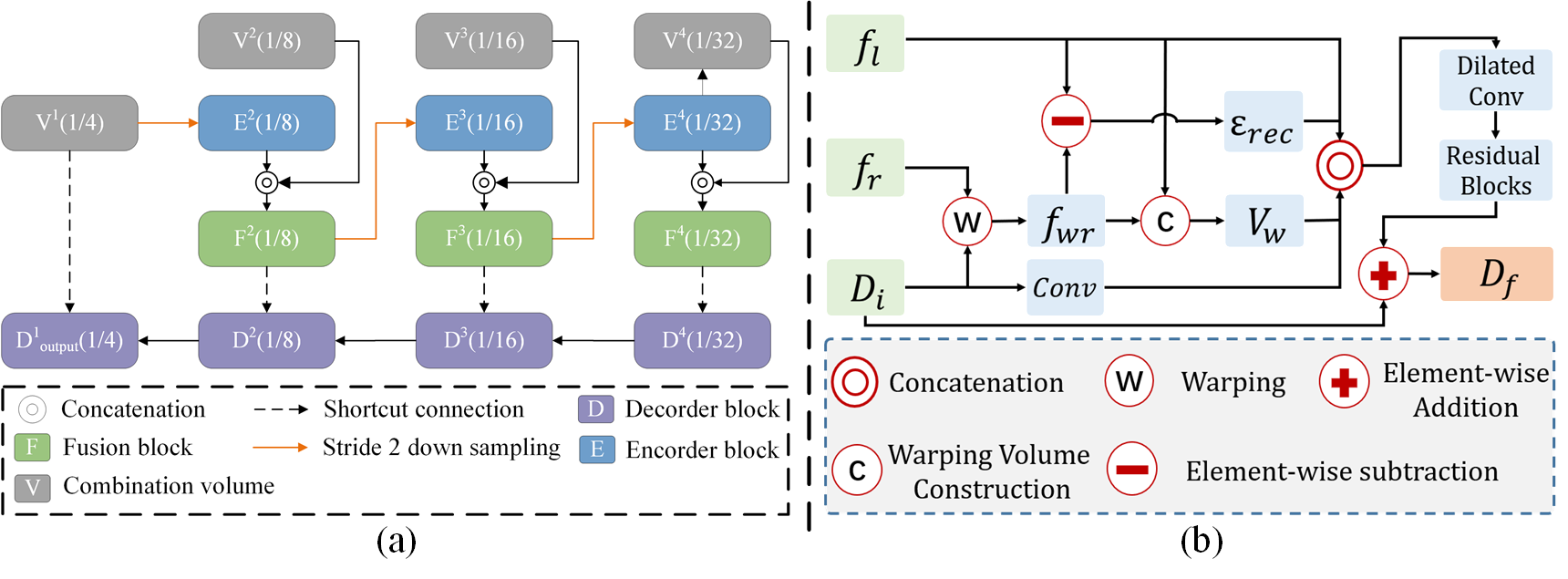} 
 \vspace{-0.15in}
 \caption{\small (a) Detailed structure of our multi-scale cost volume fusion module. (b) Detailed structure of our warping volume-based refinement network. 
 $D_{f}$ denotes the final disparity estimation.}
 \vspace{-0.25in}
 \label{fig: fusionandrefine}
\end{figure}

\subsubsection{Multi-scale Cost Volumes Fusion}
The multi-scale cost volume fusion module is shown in \figref{fig: fusionandrefine} (a), where the combination volumes, encoder blocks, fusion blocks, and decoder blocks are denoted as ${V^i},{E^i},{F^i}, {D^i}$, respectively, ($i \in \{1,2,3,4 \}$ denotes different levels). The final output fused cost volume is $D_{output}^1$. Then we use three stacked 3D hourglass networks to further process the fused cost volume and generate the initial disparity map $d_{i}$. 

\noindent\textbf{Fusion blocks.} The proposed fusion blocks have two main inputs. i) The encoder blocks, which characterize the information of higher resolution cost volume. ii) The combination volume, which directly measures the similarity between the left feature and the corresponding right feature according to a coarser disparity index. By employing the fusion blocks, we can integrate multi-scale cost volume and boost the network to evaluate the similarity of the left feature and candidate matching right feature at different scale disparity plane intervals, e.g., each disparity index represents 4 pixels interval at scale one while 32 pixels interval at scale four. Specifically, the fusion process can be formulated as:
\vspace{-0.1in}
\begin{equation}
{F^i} = \text{Conv}({V^i}||{E^i}),
    \label{eq:decoder block}
    \vspace{-0.1in}
\end{equation}
where $||$ denotes the concatenation operation at the feature axis and $\text{Conv}()$ refers to the 3D convolution layer.

\noindent\textbf{Encoder block.} Encoder block is downsampled from the previous scale fusion block by a 3D convolution with stride 2, except for ${E^1}$, which is directly downsampled from the first scale combination volume.

\noindent\textbf{Decoder blocks.} Decoder blocks comprise two main components. i) The main data flow, which continually upsamples different scale decoder blocks from ${D^4}$ to ${D^1}$. ii) The shortcut connection, which combines scale-matching fusion (encoder blocks at scale one) and decoder blocks by element-wise addition. By employing the shortcut connection, we can control the contribution of the last scale decoder block and thus balance the information flowing between upsampled decoder blocks and corresponding fusion blocks. Specifically, the decoder process can be formulated as:  
\vspace{-0.1in}
\begin{equation}
{D^i} = \left\{ \begin{array}{l}
\text{Conv}^{T}({D^{i + 1}}) + S({F^i}){\rm{  }} \quad if \quad {\rm{   }}i = 2,3,4\\
\text{Conv}^{T}({D^{i + 1}}) + S({V^i}){\rm{       }}\quad if \quad {\rm{  }}i = 1
\end{array} \right.
    \label{eq:decoder block}
    \vspace{-0.1in}
\end{equation}
Where $\text{Conv}^{T}()$ denotes the 3D transposed convolution. $S()$ refers to the shortcut connection, which is implemented by $1 \times 1 \times 1$ 3D convolution.

\subsection{Warping Volume-based Disparity Refinement}

As an essential step in typical stereo matching algorithms, disparity refinement has been widely used in deep learning-based methods. Different from previous stereo matching methods \cite{crl,mcvmsc} which learn the residual disparity value directly by the network, a multi-modal input is introduced to help our network more purposefully learn the residue. Specifically, our multi-modal input consists of the 3D warping volume, initial disparity map, left features, and reconstructed error, where the 3D warping volume is at the core. By employing the 3D warping volume, we can define a fine-grained residue searching range and alleviate the difficulty of the network to find the correct residue in an unconstrained residue searching space. Below we will describe each input in more detail.

\noindent\textbf{3D Warping Volume.} We employ the left feature and warped right feature to construct the warping volume at the last level of the pyramid. Other than the right features we used in the combination volume, we select to warp the right features according to the estimated initial disparity $D_{i}$. In this case, we can narrow down the residue searching range from initial disparity searching range $(0,D_{\max }^i)$ to a fine-grained one $(D_{i} - d_{res}, D_{i} + d_{res})$. Intuitively, the residual disparity is small. Hence, a small residue searching range $d_{res}$ is enough to correct the wrong correspondences. Specifically, the warping volume is computed as:
\vspace{-0.05in}
\begin{equation}
\begin{array}{c}
{V_{w}}(d_{res},x,y) = \frac{1}{{{N_c}}}\left\langle {{f_l}(x,y), {f_{wr}}(x - d_{res}, y)} \right\rangle, \\ \\
{f_{wr}} = \text{warping} ({f_r}, {D_{i}}),
\end{array}
\vspace{-0.051in}
\end{equation}
where $f_l$ and $f_{r}$ are upsampled from the first level feature to the original image size, $d$ denotes all residue levels in $(D_{i} - d_{res}, D_{i} + d_{res})$ and $\left\langle \:, \:\right\rangle$ represents the inner product. 

Besides, the warping operation is implemented differentially by bilinear sampling \cite{bilinearsampling}. Note that the proposed warping volume measures the similarities between left features and warped right features at each residue level which guides the network to output the optimal residual disparity with the most similarity. Moreover, we construct 3D warping volume ($H \times W \times D \times 1$) by inner product to avoid 3D convolutions which can significantly decrease the computational complexity and memory consumption.


\noindent\textbf{Reconstructed Error.} We introduce the reconstructed error to identify inaccurate regions of initial disparity estimation, which can be computed as:
\vspace{-0.07in}
\begin{eqnarray}
{\mathcal{E}_{rec}} = {f_l}(x, y) - {f_{wr}}(x, y).
\vspace{-0.12in}
\end{eqnarray}
The definition of our reconstruction error is inspired by the typical left-right consistency check, while we select to construct it at the feature level rather than the image level. By employing the reconstructed error to indicate the incorrect regions of initial disparity, our refinement network can better identify the pixels that should be further optimized. 

\noindent\textbf{Left Image Feature and Initial Disparity.} Left image features and initial disparity map are the other two inputs of our refinement network. The initial disparity map provides the network a base estimation for further optimization and the left image feature contains the context informing for residual learning. To balance the weight of multi-model input, the one-channel initial disparity map is regularized by a convolution layer to generate a 32-channel feature map.
 
\noindent\textbf{Warping Volume-based Refinement Network.} In summary, the warping volume, initial disparity map, left image features, and reconstructed error are the input of our refinement network. The detailed architecture of the refinement network is given in \figref{fig: fusionandrefine} (b). A dilated convolution \cite{dilatedconvolution} based network is employed to enlarge the receptive field which can enhance the network to give a better estimation in low-texture and occluded regions. Specifically, it has 5 convolution layers and three basic residual blocks with different dilation constants. The dilation constants are 1, 1, 2, 4, 8, 16, 1, and 1 from top to bottom.

\vspace{-0.1in}
\subsection{Loss Function}
\vspace{-0.1in}

Inspired by previous work \cite{psmnet,gwcnet}, we employ smooth $L_1$ loss function \cite{smoothl1} to train our network in an end-to-end way. For each cost volume fusion module and stacked hourglass network in cost aggregation, the same output module and soft argmin operation are used to get intermediate disparity map \cite{gwcnet}. In total, we get six disparity maps ${d_0}$, ${d_1}$, ${d_2}$, ${d_3}$, ${d_4}$, ${d_5}$ and the loss function is described as:
\vspace{-0.1in}
\begin{equation}
\mathcal{L }= \sum\limits_{j = 0}^{j = 5} {{w_j}}  \cdot  \mathcal{L}_\text{smooth-L1}({d_j} - \widehat d),
\vspace{-0.1in}
\end{equation}
$\mathcal{L}_\text{smooth-L1}$ represents the $smooth-L1$ loss and $\widehat d$ represents the ground-truth disparity and ${w_j}$ is the weight of the ${j^{th}}$ estimation of disparity map.
\vspace{-0.1in}
\section{Experimental Results} \label{sec:exps}
\vspace{-0.1in}

We evaluate our PCW-Net on various of benchmarks, including: Scene Flow \cite{dispnet}, ETH3D \cite{eth3d}, KITTI 2012\&2015 \cite{KITTI_2012,KITTI_Stereo_2015}, and Argoverse \cite{Argoverse_2_2021}. 

\subsection{Datasets} \label{subsec:datasets}

(1). \noindent \textbf{SceneFlow:} is a large synthetic dataset with 35,454 training and 4,370 test images of size $960\times 540$. It includes ``Flyingthings3D'', ``Driving'', and ``Monkaa'' with dense and accurate ground-truth for training. Here, we use the Finalpass of the Scene Flow datasets for pre-training. 
(2). \noindent \textbf{ETH3D:} is a grayscale image dataset with both indoor and outdoor scenes. The 27 training image pairs of ETH3D are employed to verify the generalization of different approaches. 
(3). \noindent \textbf{Middlebury:} is an indoor dataset with 15 training image pairs and 15 testing image pairs with full, half, and quarter resolutions. We select half-resolution training image pairs to evaluate the generalization of different approaches.
(4). \noindent \textbf{KITTI 2015 \& KITTI 2012:} are collected from the real world with a driving car. KITTI 2015 contains 200 training and 200 testing image pairs while KITTI 2012 provides 194 training and 195 testing image pairs, respectively. 
For each dataset, we select 180 image pairs from the training split for training and the rest image pairs are taken as the validation set. 
(5). \noindent \textbf{Argoverse:} is a high-resolution real-world dataset collected from a driving car. It provides 5530 training images and 1094 testing images of size $2056 \times 2464$. 
We use it to evaluate the performance of our method on high-resolution datasets, e.g., 10 times higher than KITTI.

\vspace{-0.1in}
\subsection{Implementation Details}\label{sub:implmentation}
\vspace{-0.1in}
The proposed framework is implemented using Pytorch and trained in an end-to-end manner with Adam optimizer (${\beta _{\rm{1}}} = 0.9,{\beta _2} = 0.999$). Inspired by HSM-Net \cite{hsm}, we employ asymmetric chromatic augmentation and asymmetric occlusion for data augmentation.
Moreover, we proposed a \textit{switch training strategy} to train our model for better network parameters. Specifically, it can be realized in three steps. First, the \textit{Relu} activation function is employed to train our network from scratch on the SceneFlow dataset for the first 20 epochs. We set the initial learning rate as 0.001 and down-scale it by 2 times after epoch 12, 16, 18, respectively. Then, \textit{Mish} \cite{mish} is used to prolong the pre-training process on the SceneFlow dataset for another 15 epochs. Finally, the pre-trained models are fine-tuned on KITTI 2015 and KITTI 2012 for another 400 epochs. The learning rate of this process begins at 0.001 and decreases to 0.0001 after epoch 200. Similar to other approaches, we use only the training images of KITTI 2012 for the fine-tuning process on KITTI 2012 benchmark while we merge the training images of both datasets for the KITTI 2015 benchmark. For all the experiments, the batch size is set to 4 for training on 2 NVIDIA V100 GPUs and the weights of six outputs are 0.5, 0.5, 0.5, 0.7, 1.0, and 1.3. The inherent principle of the proposed switch training strategy will be discussed in the supplementary materials.

\begin{table}[t!]
\centering
\vspace{-0.1in}
\caption{(a) Evaluation Results on the KITTI 2012\&2015 benchmark and all pixels in occluded and non-occluded areas are evaluated. (b) Evaluation Results on the Argoverse stereo benchmark. For a clear comparison, we highlight the best result in \textbf{bold} and the second-best result in \B{blue} for each column. All the metrics are the lower the better.}
    \begin{subtable}{0.43\linewidth}
      \centering
        \resizebox{1.0\textwidth}{!}{
\begin{tabular}{c|c|c|c|c|c|c|c}
\hline
\multirow{2}{*}{Methods} & \multicolumn{3}{c|}{KITTI 2015}                  & \multicolumn{4}{c}{KITTI 2012}                   \\

\cline{2-8}
                        & D1-bg         & D1-fg         & D1-all        & 2 px         & 3 px         & 4 px  & 5 px                                    \\ \hline
CSPN     \cite{cspn}                & 1.51          & \textbf{2.88} & {1.74} & \B{2.27}                    & 1.53                    & 1.19                    & 0.98                                            \\ 
GANet-deep   \cite{ganet}            & 1.48          & 3.46          & 1.81  & 2.50                    & 1.60                    & 1.23                    & 1.02                                           \\
ACFNet    \cite{acfnet}               & 1.51          & 3.80          & 1.89    & 2.35                   & 1.54                    & 1.21                   & 1.01                                            \\ 
GWCNet   \cite{gwcnet}                & 1.74          & 3.93          & 2.11   & 2.71                    & 1.70                    & 1.27                    & 1.03                                         \\ 
SSPCVNet  \cite{sspcv}               & 1.75          & 3.89          & 2.11  & 3.09                  & 1.90                   & 1.41                  & 1.14                                             \\ 
PSMNet     \cite{psmnet}              & 1.86          & 4.62          & 2.32   & 3.01                    & 1.89                    & 1.42                    & 1.15                        \\
LEAStereo \cite{lea} & \B{1.40} & \B{2.91} & \textbf{1.65}  & 2.39                  & \B{1.45}                    & \B{1.08}                    & \B{0.88}    \\ \hline

\textbf{Our PCW-Net}       & \textbf{1.37} & 3.16          & \B{1.67}   & \textbf{2.18}  & \textbf{1.37}  & \textbf{1.01}  & \textbf{0.81}                 \\ \hline
\end{tabular}
}
\vspace{-0.05in}
    \caption{}
    \end{subtable}%
    \begin{subtable}{.5\linewidth}
      \centering
        \resizebox{1\textwidth}{!}{
\begin{tabular}{c|ccc|ccc|ccc}
\hline
\multirow{2}{*}{Mehtod} & \multicolumn{3}{c|}{10 px(\%)}                                                           & \multicolumn{3}{c|}{5 px(\%)}                                          & \multicolumn{3}{c}{3 px(\%)}                                          \\ \cline{2-10} 
                        & \multicolumn{1}{c|}{all}           & \multicolumn{1}{c|}{fg}            & bg            & \multicolumn{1}{c|}{all}  & \multicolumn{1}{c|}{fg}            & bg   & \multicolumn{1}{c|}{all}  & \multicolumn{1}{c|}{fg}            & bg   \\ \hline
4Fun                    & \multicolumn{1}{c|}{\textcolor{blue}{1.79}}          & \multicolumn{1}{c|}{\textcolor{blue}{2.20}}          & \textcolor{blue}{1.62}          & \multicolumn{1}{c|}{\textcolor{blue}{3.39}} & \multicolumn{1}{c|}{\textcolor{blue}{3.07}}          & \textcolor{blue}{3.52} & \multicolumn{1}{c|}{\textcolor{blue}{6.92}} & \multicolumn{1}{c|}{4.41}          & \textcolor{blue}{7.95} \\ 
SMD-Stereo              & \multicolumn{1}{c|}{1.90}          & \multicolumn{1}{c|}{2.26}          & 1.75          & \multicolumn{1}{c|}{3.62} & \multicolumn{1}{c|}{3.15}          & 3.81 & \multicolumn{1}{c|}{7.32} & \multicolumn{1}{c|}{4.48}          & 8.49 \\ 
Cicero-stereo           & \multicolumn{1}{c|}{1.99}          & \multicolumn{1}{c|}{2.29}          & 1.87          & \multicolumn{1}{c|}{3.68} & \multicolumn{1}{c|}{3.13}          & 3.90 & \multicolumn{1}{c|}{\textbf{6.37}} & \multicolumn{1}{c|}{\textbf{4.13}}          & \textbf{7.29} \\ 
NLCANet \cite{nlcanet}          & \multicolumn{1}{c|}{2.00}          & \multicolumn{1}{c|}{2.38}          & 1.85          & \multicolumn{1}{c|}{3.69} & \multicolumn{1}{c|}{3.31}          & 3.84 & \multicolumn{1}{c|}{7.44} & \multicolumn{1}{c|}{4.60}          & 8.59 \\ 
GANet-refine \cite{ganet}           & \multicolumn{1}{c|}{2.17}          & \multicolumn{1}{c|}{2.23}          & 2.15          & \multicolumn{1}{c|}{3.73} & \multicolumn{1}{c|}{3.09}          & 3.99 & \multicolumn{1}{c|}{7.35} & \multicolumn{1}{c|}{4.43}          & 8.55 \\
CFNet \cite{cfnet}                  & \multicolumn{1}{c|}{2.38}          & \multicolumn{1}{c|}{3.79}          & 1.80          & \multicolumn{1}{c|}{4.05} & \multicolumn{1}{c|}{4.72}          & 3.78 & \multicolumn{1}{c|}{7.60} & \multicolumn{1}{c|}{6.18}          & 8.18 \\  
PSMNet \cite{psmnet}                  & \multicolumn{1}{c|}{3.05}          & \multicolumn{1}{c|}{3.81}          & 2.75          & \multicolumn{1}{c|}{4.85} & \multicolumn{1}{c|}{4.98}          & 4.79 & \multicolumn{1}{c|}{8.51} & \multicolumn{1}{c|}{6.20}          & 9.45 \\  \hline
\textbf{Our} \textbf{PCW-Net}                & \multicolumn{1}{c|}{\textbf{1.64}} & \multicolumn{1}{c|}{\textbf{1.98}} & \textbf{1.49} & \multicolumn{1}{c|}{\textbf{3.17}} & \multicolumn{1}{c|}{\textbf{2.89}} & \textbf{3.28} & \multicolumn{1}{c|}{7.05} & \multicolumn{1}{c|}{\textcolor{blue}{4.29}} & 8.18 \\ \hline
\end{tabular}
}
\vspace{-0.05in}
    \caption{}
    \end{subtable} 
\vspace{-0.2in}
\label{tab:fine-tuning_performance}
\end{table}

\begin{figure*}[!t]
    \vspace{-0.1in}
	\centering
	\small
	\tabcolsep=0.05cm
	\begin{tabular}{c c c c}
	

	\includegraphics[width=0.21\linewidth]{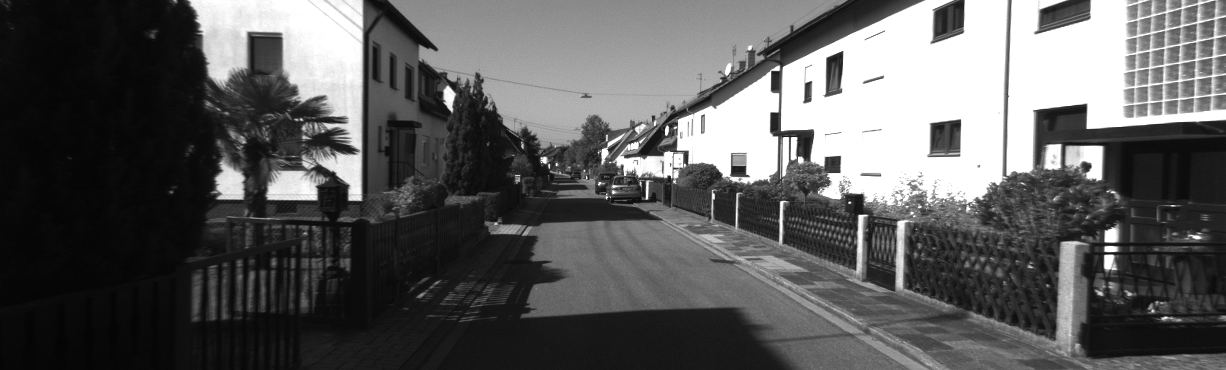}&
	\includegraphics[width=0.21\linewidth]{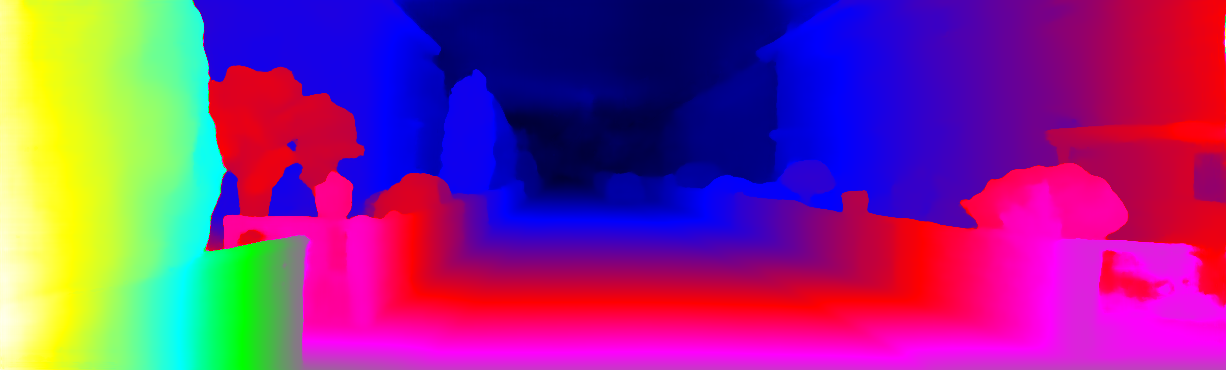}&
	\includegraphics[width=0.21\linewidth]{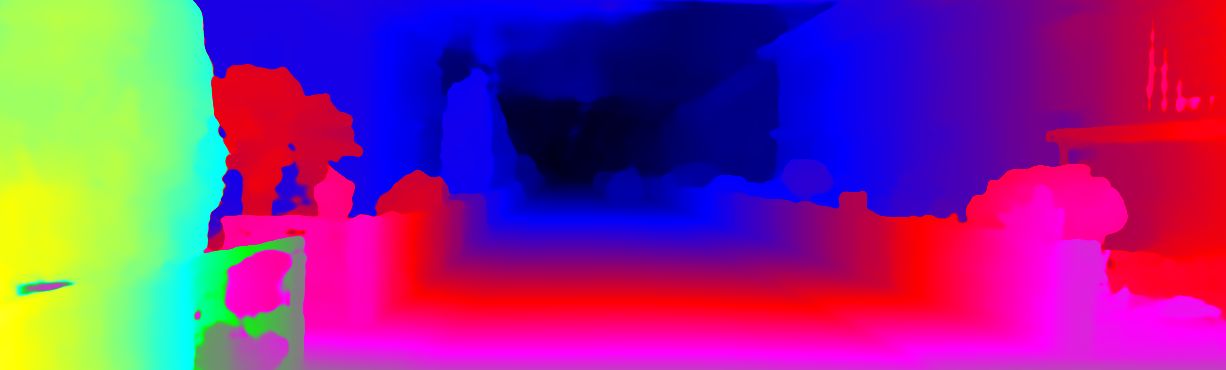}&
	\includegraphics[width=0.21\linewidth]{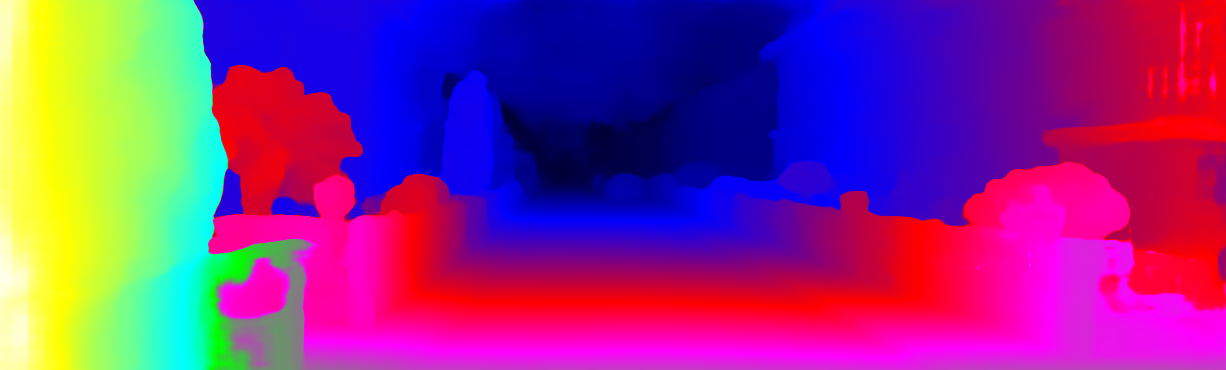}\\

	&\includegraphics[width=0.21\linewidth]{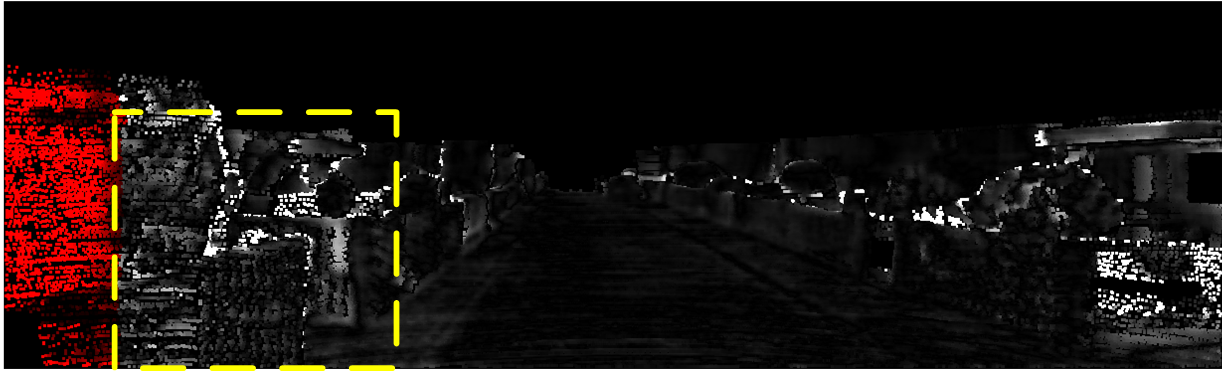}&
	\includegraphics[width=0.21\linewidth]{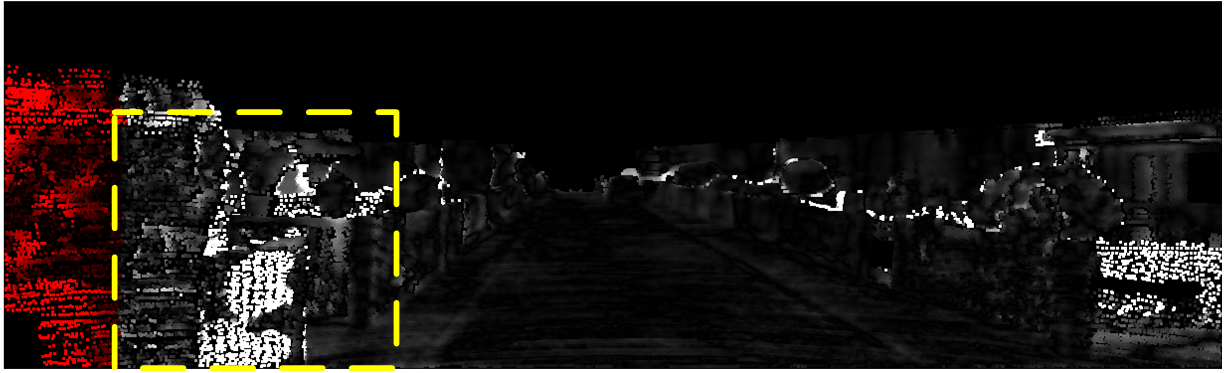}&
	\includegraphics[width=0.21\linewidth]{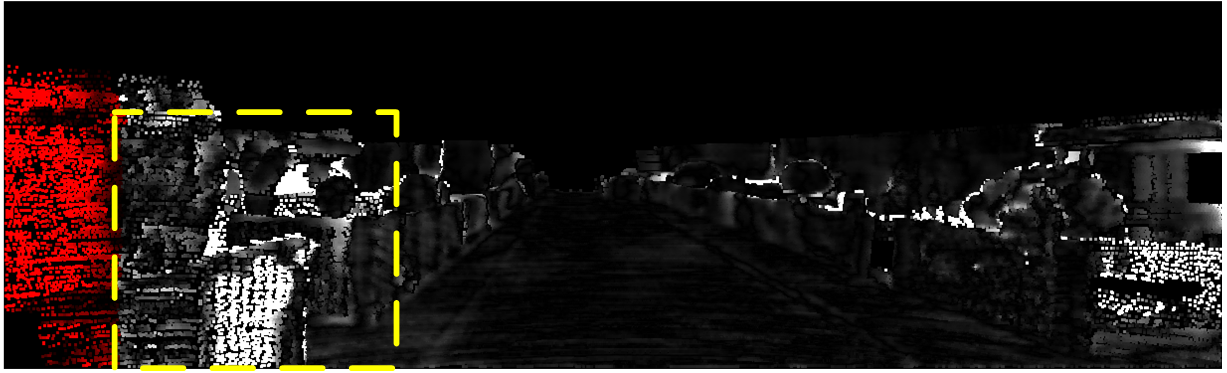}\\
	
	


	{ (a) left image} &	{(b) PCW-Net }	&  {(c) GANet-deep}	&	{(d) GWCNet}  	\\
	\end{tabular}
	
    \vspace{-0.15in}
	\caption{\small Visualization results on KITTI 2012 testset. The left panel shows the left input image of the stereo image pair, and for each example, the first row shows the predicted colorized disparity map and the second row shows the error map.}
	\vspace{-0.3in}
	\label{fig: KITTI2012}
\end{figure*}

\vspace{-0.1in}
\subsection{Fine-tuning Performance Evaluation}
\vspace{-0.1in}
In this section, we conduct experiments on various benchmarks to verify our claim in Sec. \ref{sec:intro} that the proposed method can achieve consistent SOTA fine-tuning performance on diverse real-world datasets with different proprieties. Specifically, Argoverse \cite{Argoverse_2_2021} and KITTI 2012\&2015 \cite{KITTI_2012,KITTI_Stereo_2015} are used for evaluation. Below we describe each dataset’s result in more detail.

\textbf{Results on KITTI 2012\&2015.} We train our model on the SceneFlow dataset first and then fine-tune it on the KITTI dataset. Here, we compare our fine-tuned model with other existing state-of-the-art methods. 
All results are obtained from the official KITTI evaluation website. \tabref{tab:fine-tuning_performance}(a) illustrates the comparison of the proposed method with others on the KITTI-2012. It can be shown that the proposed method achieves the best performances across all the pixels error thresholds. For the ranking criterion e.g., three-pixel-error rate, our model achieves a 1.37\% overall error rate which outperforms our base model GWCNet \cite{gwcnet} by 19.4\%. Furthermore, compared to the current best-published method LEAStereo \cite{lea}, our method can also achieve a 5.5\% error reduction on the overall three-pixel-error rate.

The comparison with other state-of-the-art approaches on the KITTI-2015 benchmark is given in \tabref{tab:fine-tuning_performance}(a). From this table, generally, we can easily find that the proposed method achieves 1 first-place and 1 second-places among all the three categories. Specifically, our method achieves a 1.67\% overall three-pixel-error rate, which surpasses the base model GWCNet by 20.85\%. Compared to LEAStereo \cite{lea}, we can obtain very similar results, especially for the ranking criterion ``D1-all'' category (1.65 vs 1.67).


Qualitative comparison results on the KITTI 2012 benchmark are shown in Fig.~\ref{fig: KITTI2012}, and we can see that our method shows significant improvement in ill-posed regions and fence regions (see dash boxes in the picture). The visualization results further support our claim that employing multi-scale cost volumes can guide the network to learn the affiliation between an object and its sub-region, thus promoting the estimation of the textureless region and repeated pattern. More qualitative results are given in the supplementary materials.

\textbf{Results on Argoverse.} Argoverse is a high-resolution real-world dataset collected from a driving car. In comparison to KITTI, it has 10 times the resolution and 16 times as many training frames, making it a more robust and challenging dataset. Similar to the KITTI, we train our model on the SceneFlow dataset first and then fine-tuning it on the Argoverse dataset. Here, we compare our fine-tuned model with other existing state-of-the-art methods in \tabref{tab:fine-tuning_performance}(b). All results are obtained from the official Argoverse evaluation website\footnote{\href{https://eval.ai/web/challenges/challenge-page/917/leaderboard/2412}{https://eval.ai/web/challenges/challenge-page/917/leaderboard/2412}}.
To be clear, \textit{the 10-pixel error} is taken as the \textit{official evaluation metric} in this benchmark due to its high image resolution.
From this table, we can easily find that existing state-of-the-art stereo matching methods \cite{ganet,psmnet,nlcanet,cfnet} cannot achieve consistent finetuning performance on the Argoverse dataset. This is likely caused by the different proprieties between KITTI and Argoverse, e.g., high-resolution vs low-resolution and large-scale dataset vs small-scale dataset. Instead, as shown in \tabref{tab:fine-tuning_performance}(b), we can easily find that the proposed method achieves 6 first places among all the nine categories, which further verifies our claim that the proposed method can achieve consistent performance on diverse datasets. We attribute this result to the proposed multi-scale cost volume fusion module, which can cover multi-scale receptive fields and boost the network to see different scale regions of the original image. Such an operation is well suited for both low-resolution and high-resolution images. Specifically, our method outperforms state-of-the-art approaches on overall ten-pixel-error and five-pixel-error rates with 1.64\% and 3.17\%. Cicero-stereo is the best method on the three-pixel-error rate and our method can achieve comparable results with it, especially for the ranking criterion ``fg'' category (4.29\% vs 4.13\%). Note that the evaluation images in Argoverse Dataset are with high resolution ($2056 \times 2464$). Thus, ten-pixel-error and five-pixel-error are the main evaluation metrics. All in all, our method ranks $1^{st}$ on the Argoverse leaderboard and sets a new SOTA performance.

\begin{table}[t!]
\vspace{-0.1in}
\caption{\small Cross-domain generalization evaluation on four real datasets. For a fair comparison, all methods are only trained on the SceneFlow training set and tested on four real datasets. We highlight the best result in \textbf{bold} and the second-best result in \B{blue} for each column. All the metrics are the lower the better. Half resolution training sets of Middlebury is employed for evaluation.}
\centering
\resizebox{0.7\textwidth}{!}{
\begin{tabular}{c|c|c|c|c|c}
\hline
Method    & \begin{tabular}[c]{@{}c@{}}KITTI2012\\ D1\_all(\%)\end{tabular} & \begin{tabular}[c]{@{}c@{}}KITTI2015\\ D1\_all(\%)\end{tabular} & \begin{tabular}[c]{@{}c@{}}Middlebury(half)\\ bad 2.0(\%)\end{tabular} & \begin{tabular}[c]{@{}c@{}}ETH3D\\ bad 1.0(\%)\end{tabular} & time (s)\\ \hline
HD\textasciicircum{}3 & 23.6  & 26.5          & 37.9  & 54.2  &\textbf{0.14}\\

PSMNet   & 15.1                                                            & 16.3                                                            & 25.1                                                            & 23.8   &0.41                                                      \\ 
GWCNet    & 12.0                                                            & 12.2                                                            & 24.1                                                             & 11.0     &0.32                                                    \\  
GANet    & 10.1                                                            & 11.7                                                            &  20.3                                                            & 14.1         &1.8                                               \\ 
DSMNet   & 6.2                                                             & 6.5                                                             & \textbf{13.8}                                                   & 6.2             &1.5                                            \\ 
CFNet     &\textbf{\B{4.7}}                                                    & \textbf{\B{5.8}}                                                    & 19.5                                                            & \textbf{\B{5.8}}                  &0.22                             \\ \hline
\textbf{Our PC-Net}  & 4.5     & 5.8   & 19.00  & 5.4       &0.33    \\ 
\textbf{Our PCW-Net} & \textbf{4.2}   & \textbf{5.6}  & \textbf{\B{15.77}} & \textbf{5.2}      &0.44                            \\ \hline
\end{tabular}
}
\label{tab:generalization}
\vspace{-0.15in}
\end{table}

 \begin{figure*}[!t]
 	\centering
 	\small
 	\tabcolsep=0.05cm
 	\begin{tabular}{c c c c}
 	
 	\includegraphics[width=0.21\linewidth]{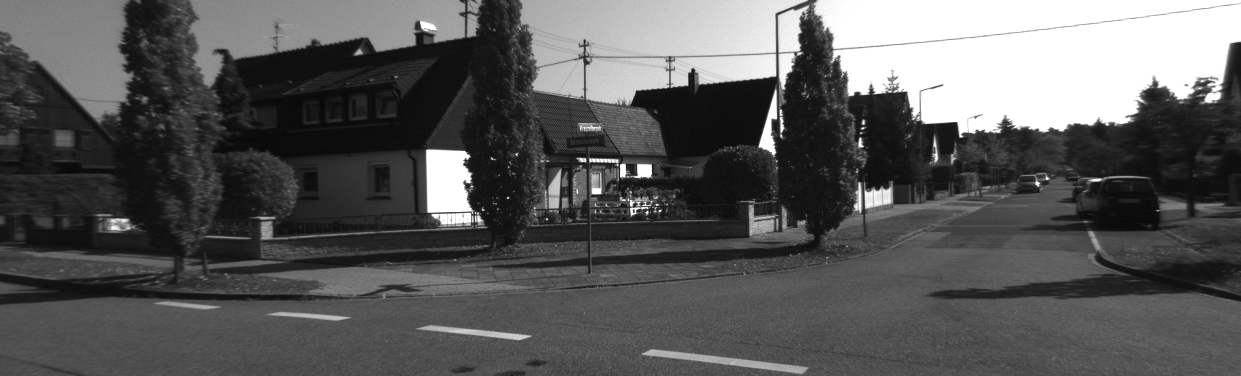}&
 	\includegraphics[width=0.21\linewidth]{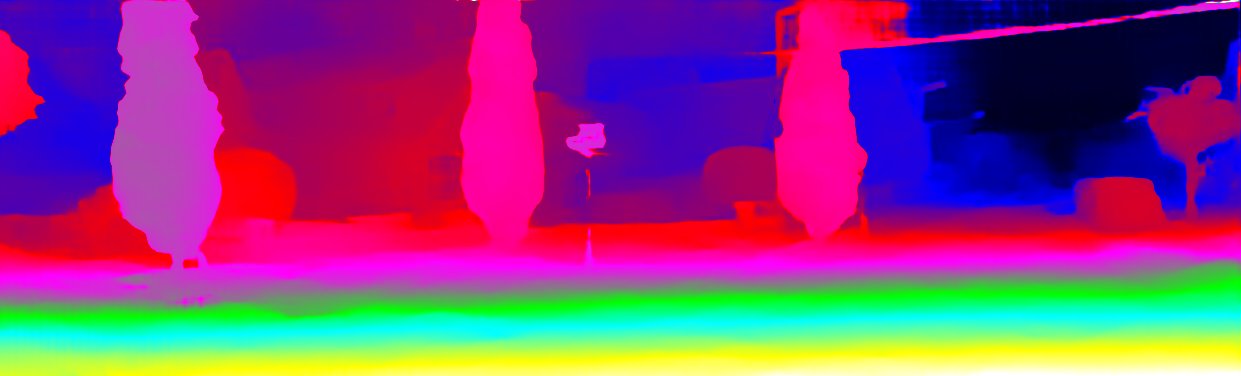}&
 	\includegraphics[width=0.21\linewidth]{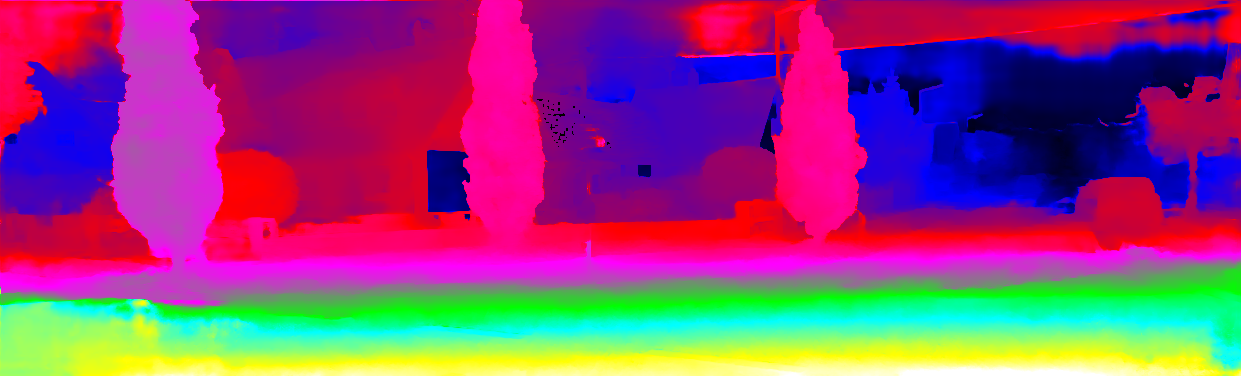}&
 	\includegraphics[width=0.21\linewidth]{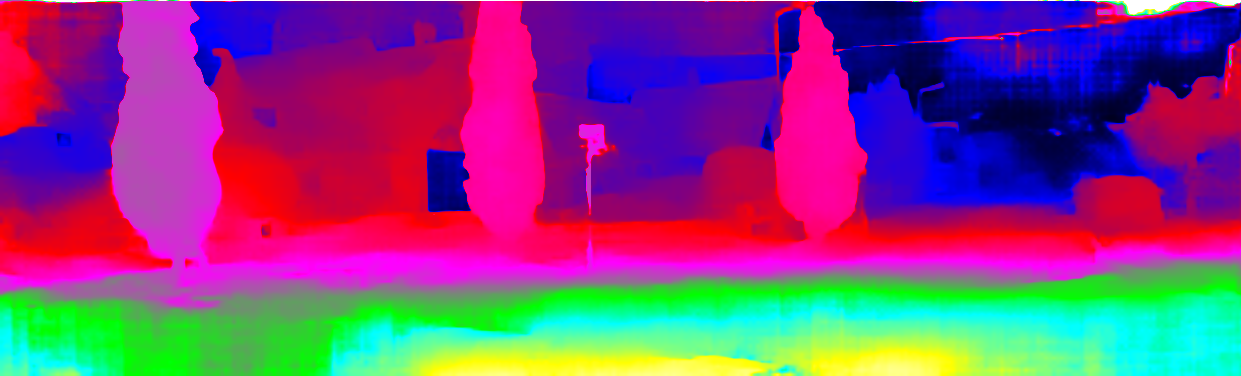}\\
	
 	&\includegraphics[width=0.21\linewidth]{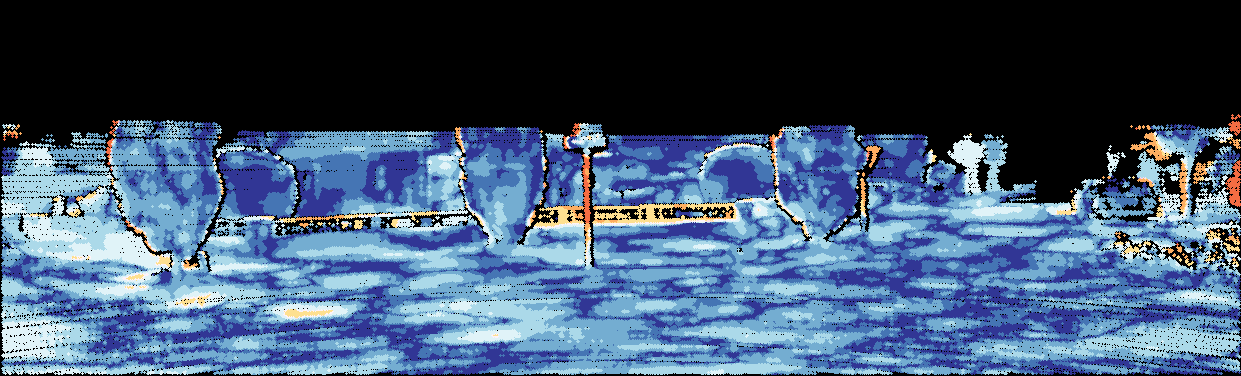}&
 	\includegraphics[width=0.21\linewidth]{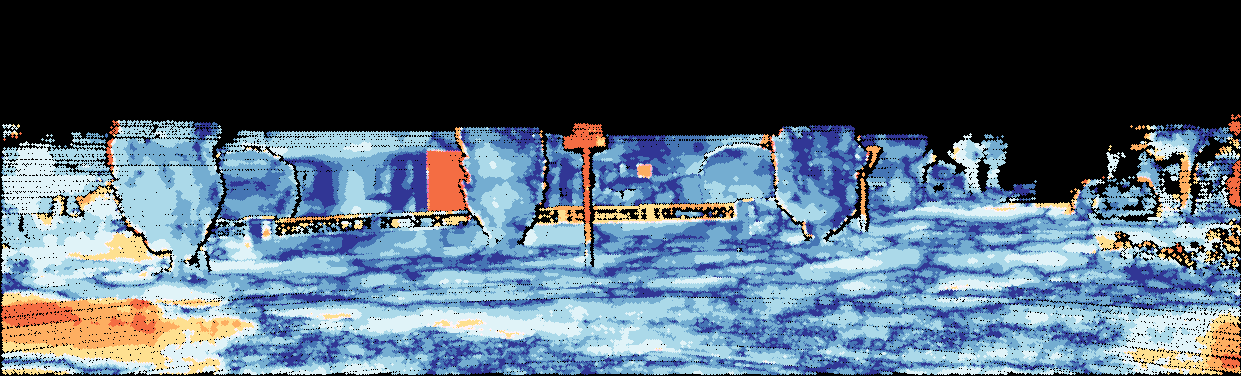}&
 	\includegraphics[width=0.21\linewidth]{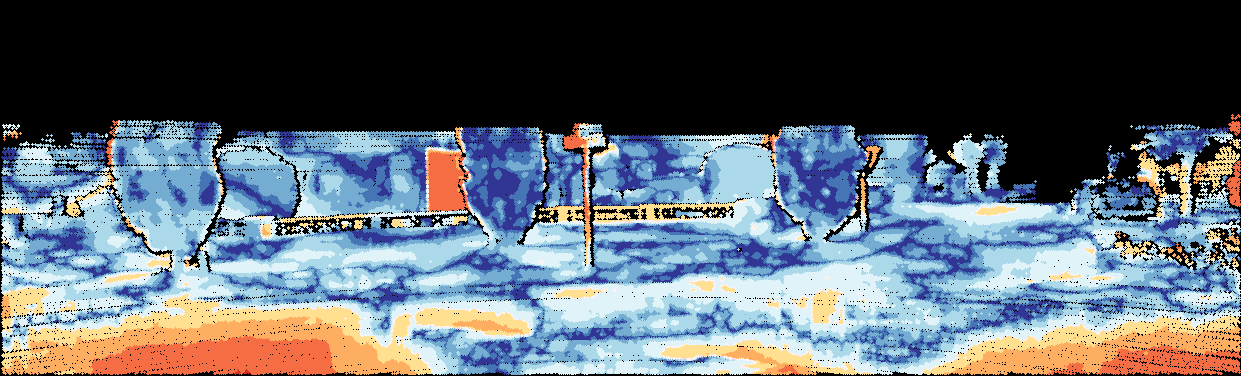}\\

 	{ \footnotesize (a) left image} &	{\footnotesize (b) PCW-Net }	&  {\footnotesize (c) GANet-deep}	&	{\footnotesize (d) GWCNet}  	\\
 	\end{tabular}
	    \vspace{-0.15in}
 	\caption{\small Cross-domain generalization comparison on KITTI2012 trainset. All methods are trained on the synthetic dataset and tested on KITTI2012 trainsets. The left panel shows the left input image of stereo image pairs, and for each example, the first row shows the predicted colorized disparity map and the second row shows the error map.}
 	\vspace{-0.2in}
 	\label{fig: generalization}
 \end{figure*}

\vspace{-0.1in}
\subsection{Cross-Domain Generalization Evaluation} \label{subsec:corss_domain}
In this section, we conduct experiments to verify our claim in Sec. \ref{sec:intro} that the proposed PCWNet can achieve strong cross-domain generalization. Specifically, we design an experiment by training the model on the synthetic data only and testing it on four real datasets such as KITTI 2012, KITTI 2015, ETH3D, and Middlebury. To make a fair comparison, all the methods are trained only on the Scene Flow dataset (without any other synthetic or real data will be used, i.e., Carla \cite{dsmnet}). The comparison with other approaches is given in \tabref{tab:generalization}. From this table, we can find that our method outperforms the baseline model \textit{gwcnet} on all four datasets with a large margin. Compared to the second-best method CFNet\cite{cfnet}, our proposed PCNet (refers to the network without the warping volume based disparity refinement) has achieved comparable performance and the proposed PCWNet can further surpass it on all four datasets. Specifically, the error rate on KITTI 2012, KITTI 2015, ETH3D, and Middlebury has been decreased by 10.64\%, 3.45\%, 10.34\%, and 19.13\%, respectively compared to CFNet. Most importantly, both CFNet and DSMNet are specially designed for cross-domain generalization and will make a significant compromise on finetuning performance, e.g., the D1\_all error rate of CFNet \cite{cfnet} on the KITTI 2015 benchmark is 1.88\%, which is 10.11\% higher than ours. In summary, the comparison between these domain-generalization methods further shows that our PCW-Net can make a good balance between performance and generalization.

In addition, we compare the generalization results of our method with some state-of-the-art methods in Fig.~\ref{fig: generalization}. From this figure, we can clearly see that most existing dataset-specific methods \cite{gwcnet,ganet} generalize poorly to unseen real scenes while our method can correct most errors and generate a reasonable result. More qualitative results on other datasets will be given in the supplementary materials.

\vspace{-0.1in}
\subsection{Ablation Studies} \label{sec:ablation_studies}

To verify the effectiveness of different modules, we set a series of experiments in this section. For efficient evaluation, only the KITTI 2015 dataset (without pre-training from Scene Flow) has been used for training and evaluation. Generally, four types of experiments have been executed here.

\noindent\textbf{Multi-scale cost volume fusion.} The proposed multi-scale cost volume fusion module consists of the combination volumes, encoder blocks, fusion blocks, and decoder blocks. Here, we verify the impact of removing the fusion blocks, which means that the multi-scale combination volume information is ignored. As shown in the $Multi-scale$ $Cost$ $Volume$ $Fusion$ section of Tab. \ref{tab: abstudy_refinement}, the D1\_all error rate increase from 1.97\% (D+E+F(ours)) to 2.09\% (D+E) after removing the fusion blocks, which further verifies the necessity of including multi-scale information.

\noindent\textbf{Cost volume construction.} The proposed combination volume consists of concatenation volume and group-wise correlation volume. Here, We test the impact of using different cost volumes. As shown in \tabref{tab: abstudy_refinement}, the proposed combination volume achieves the best result. Moreover, the performance of combination volume without the normalization layer $\delta$ is even worse than the usage of single cost volume. Thus, it’s essential to add this layer to make the two cost volumes share the same data distribution.

\noindent\textbf{Multi-modal input evaluation.} In the disparity refinement module, we employ multi-modal input to help our network learn the residue more purposefully. Here, we test the impact of each input individually.
As shown in the \emph{Multi-modal} $input$ section of Tab. \ref{tab: abstudy_refinement}, each input is indispensable and the 3D warping volume is at the core. Specifically, the improvements of each part are: 0.29\% for $V_w$, 0.08\% for ${\mathcal{E}_{rec}}$, 0.02\% for $f_i$ and $D_i$, respectively. The result verifies all the multi-modal inputs work positively to improve the performance and compared with other inputs, the 3D warping volume $V_w$ achieves the largest gain.



\noindent\textbf{Model Generalization.} Moreover, to further verify the generalization of the proposed method, we conduct two more ablation studies.
In this setting, all the frameworks are trained on the SceneFlow dataset and evaluated on the SceneFlow testing set and KITTI 2015 training set without finetuning. The comparison results are given in \tabref{tab: ablation_generalization} (a). From the table, we can find that the proposed multi-scale cost volume fusion(MSCVF) and warping volume based disparity refinement (WVBDF) can both promote the generalization ability on KITTI as well as finetuning performance on SceneFlow. The error on the KITTI dataset has been decreased from 6.18\% to 5.55\%. Moreover, we further analyze the effect of each module on generalization in \tabref{tab: ablation_generalization} (b). As shown, each module works positively for better generalization, and the multi-scale cost volume fusion module (MSCVF) is at the core, which contributes 68.19\%, 68.54\%, 57.65\% error reduction on KITTI2012, KITTI2015, and ETH3D, respectively.

\renewcommand\arraystretch{1.2}
\begin{table}[!t]
\vspace{-0.1in}
\caption{\footnotesize Ablation Study of the proposed method on the KITTI2015 dataset. $V_{w}$, ${\mathcal{E}_{rec}}$, $f_l$, $D_{i}$ denote the 3D warping volume, reconstructed Error, left features, and initial disparity map, respectively. D, E, and F represent decoder blocks, encoder blocks, and fusion blocks, respectively. D1\_all is used for evaluation (the lower the better). We test a component of our method individually in each section of the table and the approach which is used in our final model is underlined.}
\centering
\resizebox{0.6\textwidth}{!}{
\begin{tabular}{c|c|c}
\hline
Experiment                                      & Method                          & \begin{tabular}[c]{@{}c@{}}KITTI\\ D1\_all\end{tabular} \\ \hline
\multirow{2}{*}{Multi-scale Cost Volume Fusion} 
& D+E  & 2.09 \\ \cline{2-3} 
& {\ul D+E+F (ours)}     & \textbf{1.97} \\ \hline
\multirow{4}{*}{Cost Volume} & concatenation volume               & 2.04 \\ \cline{2-3} 
&  group-wise correlation volume   & 2.13       \\ \cline{2-3} 
&  combination volume without $C_r$   & 2.14  \\ \cline{2-3} 
& {\ul combination volume (ours)}  & \textbf{1.97}   \\ \hline
\multirow{4}{*}{Multi-modal input} & Multi-modal input without $V_{w}$               & 2.26 \\ \cline{2-3} 
& Multi-modal input without ${\mathcal{E}_{rec}}$   & 2.05       \\ \cline{2-3} 
& Multi-modal input without $f_l$ and $D_{i}$  & 1.99  \\ \cline{2-3} 
& {\ul Multi-modal input (ours)}  & \textbf{1.97}   \\ \hline
\end{tabular}
}
\label{tab: abstudy_refinement}
\renewcommand\arraystretch{1}

\caption{\footnotesize (a) Ablation study of model generalization. (b) Sub-module generalization analysis on three real datasets. All methods are only trained on the synthetic dataset and tested on three real datasets. MSCVF and WVBDF denote the multi-scale cost volume fusion module and warping volume based disparity refinement module, respectively. Raw disparity refers to the disparity estimation result before cost volume fusion.}
\vspace{-0.15in}
\centering
    \begin{subtable}{0.5\linewidth}
      \centering
        \resizebox{0.95\textwidth}{!}{
        \begin{tabular}{c|c|c}
\hline
\multirow{2}{*}{Method}  & SceneFlow       & \begin{tabular}[c]{@{}c@{}}KITTI 2015 (w/o finetuning)\end{tabular} \\  
                                 & EPE (px)         & D1\_all (\%)                                                              \\ \hline
no WVBDF + MSCVF                             & 0.8578          & 6.18                                                                     \\ 
no WVBDF                            & 0.8387          & 5.81                                                                     \\ 
PCWNet                          & \textbf{0.7868} & \textbf{5.55}                                                            \\ \hline

\end{tabular}

}
\vspace{-0.1in}
    \caption{}
    \end{subtable}%
    \begin{subtable}{0.5\linewidth}
      \centering
        \resizebox{0.8\textwidth}{!}{
\begin{tabular}{r|c|c|c}
\hline
Different operations    & \begin{tabular}[c]{@{}c@{}}KITTI 2012\\ D1\_all    (\%)\end{tabular} & \begin{tabular}[c]{@{}c@{}}KITTI 2015\\ D1\_all  (\%)\end{tabular} & \begin{tabular}[c]{@{}c@{}}ETH3D\\ bad  1.0 (\%)\end{tabular} \\ \hline
raw disparity & 8.60  & 8.57           & 16.44                              \\ 
MSCVF   & 5.62 (\textbf{-2.98})   & 6.5  (\textbf{-2.07})  & 9.96 (\textbf{-6.48})     \\ 
Stacked hourglass  & 4.49 (-1.13)  & 5.84 (-0.66)           & 6.57  (-3.39)           \\ 
WVBDF   & 4.23 (-0.26)   & 5.55 (-0.29)           & 5.2 (-1.37)           \\ \hline 
\end{tabular}
}
\vspace{-0.1in}
    \caption{}
    \end{subtable} 
\label{tab: ablation_generalization}
\vspace{-0.35in}
\end{table}

\vspace{-0.1in}
\section{Conclusion} \label{sec:conclusion}
\vspace{-0.1in}
In this paper, we have proposed a pyramid combination and warping cost volume based network, $i.e.$, PCW-Net, for accurate and robust stereo matching. Our pyramid cost volume can be divided into two parts. First, we construct combination volumes on the upper levels of the pyramid and develop a cost volume fusion module to integrate them for initial disparity estimation. Second, we construct the warping volume on the last level of the pyramid and employ it to refine the initial disparity. 
Experimental results show the superiority of PCW-Net across a diverse range of datasets. Specifically, PCW-Net achieves state-of-the-art performance and strong cross-domain generalization at the same time. 
In the future, we plan to explore our cost volume representation to other dense matching problems such as multi-view stereo. 

\noindent\textbf{Acknowledgements} {This research was supported in part by National Key Research and Development Program of China (2018AAA0102803) and NSFC (61871325). 
}
\bibliographystyle{splncs04.bst}
\bibliography{egbib}
\end{document}